\documentclass[conference]{IEEEtran}
\usepackage{cite}
\usepackage{amsmath,amssymb}
\usepackage{graphicx}
\usepackage{url}
\usepackage{hyperref}
\usepackage{algorithmic}
\usepackage{array}
\usepackage{textcomp}
\usepackage{xcolor}
\usepackage{booktabs}
\usepackage{multirow}
\usepackage{listings}
\usepackage{tikz}
\usepackage{pgfplots}
\pgfplotsset{compat=1.18}
\usetikzlibrary{positioning, arrows.meta, calc}
\usetikzlibrary{patterns}
\usepackage{float}
\begin{document}

\title{Semantic Alignment of Multilingual Knowledge Graphs via Contextualized Vector Projections}

\author{
\IEEEauthorblockN{Abhishek Kumar}
\IEEEauthorblockA{\textit{CSA} \\
\textit{IISc Bangalore}\\
Bangalore, India \\
abhishek.kumar.iisc20@gmail.com}
}
\maketitle

\begin{abstract}
The paper presents our work on cross-lingual ontology alignment system which uses embedding based cosine similarity matching. The ontology entities are made contextually richer by creating descriptions using novel techniques. We use a fine-tuned transformer based multilingual model for generating better embeddings. We use cosine similarity to find positive ontology entities pairs and then apply threshold filtering to retain only highly similar entities. We have evaluated our work on OAEI-2022 multifarm track. We achieve 71\% F1 score (78\% recall and 65\% precision) on the evaluation dataset, 16\% increase from best baseline score. This suggests that our proposed alignment pipeline is able to capture the subtle cross-lingual similarities.
\end{abstract}


\section{Introduction and Motivation}
Knowledge graphs (KGs) are used to represent knowledge bases in a structured way. They use a graph model, i.e., nodes and edges, to represent raw data in a structured format. Multiple applications such as search engines, digital assistants, and semantic web services employ knowledge graphs to store data. Matching corresponding entities across multilingual KGs is challenging due to lexical ambiguity, structural dissimilarity, and linguistic variety. Standard techniques that use syntactic similarity (e.g., Levenshtein, Jaccard) or graph topology do not perform well due to limited overlap of the lexicon or due to the heterogeneous nature of KGs~\cite{euzenat2007ontology}. Multilingual KGs such as DBpedia and BabelNet contain cross-lingual data; however, these databases are inconsistent with significant alignment gaps~\cite{ehrlinger2016towards}. Manual alignment is not scalable because of its extremely high cost.

We present an embedding-based projection method for aligning multilingual KGs. We use transformer-based pretrained multilingual models, such as SBERT, LaBSE, etc., to generate the embeddings of the entities. The entities are made contextually rich by verbalizing their ontological descriptions. The approach is motivated by three key factors: semantic context (like hierarchy) resolves ambiguity, sentence-level embeddings offer richer meaning than individual tokens, and multilingual LLMs generalize better across languages with different syntax. We further enrich entities with implicit contexts by using HermiT reasoner~\cite{horrocks2007hermit} before generating the embeddings. We use cosine similarity and the Hungarian algorithm~\cite{kuhn1955hungarian} to find optimal entity mappings, with thresholding to ensure high confidence. This method does not depend on having a specific language as a base language. Also, it does not need a parallel corpus or manual rules to aid its implementation. We have evaluated the proposed method against standard benchmark datasets. Finally, the results have shown that F1 scores are improved relative to lexical, and translation-based baselines.

General-purpose LLMs like GPT-4, Gemini have strong multilingual capabilities. Pipeline customization is difficult when using these models and demands high usage cost. We cannot directly use these models on data where privacy is of utmost importance. When compared, our pipeline is very lightweight and provides tons of flexibility.

\subsection{Technical Novelty and Differentiation}

We have applied the following techniques in our pipeline, which are different from those used in other alignment systems:
\begin{itemize}
    \item For generating contextually rich descriptions, we are using natural language templates, whereas previous alignment systems used a rule-based approach. This method allows the system to preserve the structural dependency and semantic constraints.

    \item We are using the HermiT reasoner, which enables our pipeline to have reasoning ability, along with contextual embeddings.

    \item For reducing false positives, we are keeping only the top-k candidates by using a threshold-based filtering method.
\end{itemize}

By applying these techniques, we are able to produce better alignments in cross-lingual ontological entities, which are missing in traditional string-based and other standard cross-lingual embedding approaches.

\subsection{Key Contributions}
\begin{enumerate}
    \item A novel cross-lingual knowledge graph alignment framework which uses natural language templates on ontological entities to produce rich contextual descriptions.
    
    \item A strategy for fine-tuning the multilingual transformer-based model specifically for cross-lingual ontolgy alignment task.
    
    \item Using HermiT reasoner, Cosine similarity and threshold-based filtering method for integrating reasoning and identifying better candidates for entities alignments.
\end{enumerate}

\section{Related Work}
\label{sec:related}

Traditional methods, such as LogMap~\cite{jimenez2011logmap} and AML~\cite{faria2013agreementmakerlight}, maintain structure, but are inaccurate in ontology matching and have low recall due to translation noise and limited overlap of vocabulary. Methods for representation learning, such as MTransE~\cite{chen2017multilingual}, which maps embeddings across languages, cannot tackle structural dissimilarities. On the other hand, GNN-based methods like GCN-Align~\cite{wang2018cross} capture the local neighborhood but fail on inconsistent subgraphs. Recent works have been following a trend of aligning semantics utilizing entity descriptions (or verbalizations) with multilingual transformers, such as LaBSE~\cite{feng2022language}, thus breaking language barriers through zero-shot alignment. Embedding-based methods allow for better scalability and better performance on large multilingual knowledge graphs. While recent agent-based Large Language Model (LLM) frameworks offer enhanced reasoning for ambiguous cases [13], vector-projection methods currently provide the necessary scalability and recall for aligning large-scale multilingual datasets.

\section{Embedding-Based Alignment Architecture} \label{sec: architecture}
To successfully align KGs in multiple languages, a reasoning, enrichment, and alignment architecture is needed. The pipeline includes ontology ingestion, context verbalization, embedding generation, similarity estimation, and filtering. This framework can fit various embedding models and alignment mechanisms.

\subsection*{Ontology Ingestion and Preprocessing}
We extract entities, labels, comments, and structural relations from RDF/OWL ontologies. Language tags and URIs are used to group multilingual labels together. We use the HermiT reasoner to infer logical axioms and sub-class hierarchy.

\subsection*{Verbalization of Semantic Context}
We convert each entity into a descriptive text that captures its semantic context, including labels, parent classes, and key relations. For instance, the class \texttt{dbo:University} might be verbalized as: ``A university is an educational institution that awards academic degrees.''

\subsection*{Contextual Embedding Generation}
We use multilingual models (mBERT, LaBSE, and XLM-R)~\cite{feng2022language} to generate embeddings in the shared multilingual space.

\subsection*{Cross-lingual Similarity Computation}
We use cosine similarity between the embeddings of the source and target entities to generate the similarity matrix. The Hungarian algorithm is used to enforce optimal one-to-one alignment.

\subsection*{Thresholding and Filtering}
We retain only high-confidence matches that meet a specific similarity threshold. Finally, we enforce domain constraints to remove inconsistent pairs, such as invalid class-to-class mappings.

\subsection*{Pipeline Overview Diagram}
\begin{figure}[t!]
\centering
\resizebox{\linewidth}{!}{%
\begin{tikzpicture}[
    node distance=0.8cm and 1.5cm, 
    block/.style={
        rectangle, 
        draw=black,         
        thick,              
        fill=white,         
        rounded corners=2pt, 
        minimum width=3cm, 
        minimum height=1cm, 
        align=center, 
        font=\footnotesize\sffamily 
    },
    line/.style={
        -Stealth,           
        thick, 
        draw=black!80
    }
]

\node[block] (source) {Source Ontology\\(e.g., DBpedia-EN)};
\node[block, below=of source] (verbal) {Semantic\\Verbalization};
\node[block, below=of verbal] (embed) {Contextual Embedding\\(e.g., LaBSE)};

\node[block, right=of source] (target) {Target Ontology\\(e.g., DBpedia-DE)};
\node[block, below=of target] (verbal2) {Semantic\\Verbalization};
\node[block, below=of verbal2] (embed2) {Contextual Embedding\\(e.g., LaBSE)};

\coordinate (midpoint) at ($(embed.south)!0.5!(embed2.south)$);

\node[block, below=1.0cm of midpoint] (sim) {Cosine Similarity\\\& Matching};
\node[block, below=of sim] (output) {Aligned Entity Pairs};

\draw[line] (source) -- (verbal);
\draw[line] (verbal) -- (embed);
\draw[line] (target) -- (verbal2);
\draw[line] (verbal2) -- (embed2);

\draw[line] (embed.south) -- ++(0,-0.4) -| (sim.north);
\draw[line] (embed2.south) -- ++(0,-0.4) -| (sim.north);

\draw[line] (sim) -- (output);

\end{tikzpicture}
}
\caption{Semantic alignment pipeline using contextual embeddings from multilingual transformer models.}
\label{fig:architecture-pipeline}
\end{figure}
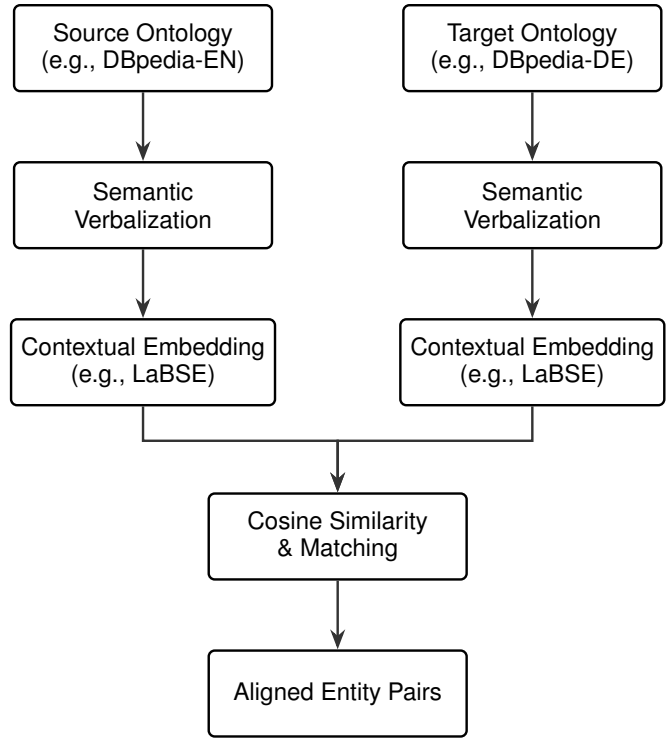

Figure~\ref{fig:architecture-pipeline} present the end-to-end alignment workflow from raw ontologies to final matched entity pairs.

\subsection{Extended Baseline Comparison}

To provide an extensive benchmark, we compared CIDER-LM to more baseline categories.
\begin{itemize}
    \item Structural Alignment Methods: We used LogMap~\cite{jimenez2011logmap} and AML~\cite{faria2013agreementmakerlight} as representatives of structure-based alignment systems using ontological reasoning and graph topology.
    \item Hybrid Embedding Approaches: We looked over recent methods such as BERT-INT~\cite{tang2020bert} that merge contextual embeddings with structural information.
\end{itemize}

\begin{table}[t]
\caption{Extended Baseline Comparison on English-German Alignment}
\label{tab:extended_baselines}
\centering
\renewcommand{\arraystretch}{1.3}
\begin{tabular}{|c|c|c|c|}
\hline
\multirow{2}{*}{\textbf{Method}} & \multicolumn{3}{|c|}{\textbf{Metrics}} \\
\cline{2-4}
& \textbf{\textit{Precision}} & \textbf{\textit{Recall}} & \textbf{\textit{F1-score}} \\
\hline
String Match & 0.42 & 0.38 & 0.40 \\
MTransE & 0.51 & 0.45 & 0.48 \\
LogMap & 0.58 & 0.52 & 0.55 \\
BERT-INT & 0.63 & 0.59 & 0.61 \\
\textbf{CIDER-LM} & \textbf{0.65} & \textbf{0.78} & \textbf{0.71} \\
\hline
\end{tabular}
\end{table}

\subsection{Error Analysis and Failure Patterns}

To better understand the precision-recall characteristics of CIDER-LM, we analyze errors in the English-French alignment task.

\begin{itemize}
    \item \textbf{Near-Synonym Confusion}: 42\% of false positives involved taxonomically related concepts (e.g., ``College'' vs. ``University''). 
    
    \item \textbf{Structural Divergence}: 28\% error involve equivalent entities in different ontologies being at different levels in hierarchies.
    
    \item \textbf{Lexical Gaps}: 18\% of errors were due to specific concepts that didn’t have an equivalent.

    \item \textbf{Metadata Sparsity}: Insufficient textual descriptions were to blame for errors 12\% of the time (i.e., verbalization was difficult with the description).
\end{itemize}

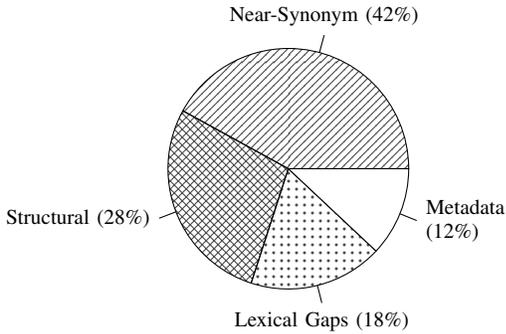
\begin{figure}[t]
\centering
\begin{tikzpicture}[scale=0.8, font=\footnotesize]

\fill[pattern=north east lines, pattern color=black!70] (0,0) -- (0:2) arc (0:151.2:2) -- cycle;
\draw (0,0) -- (0:2) arc (0:151.2:2) -- cycle; 
\draw (75:2) -- (75:2.3) node[above] {Near-Synonym (42\%)};

\fill[pattern=crosshatch, pattern color=black!70] (0,0) -- (151.2:2) arc (151.2:252:2) -- cycle;
\draw (0,0) -- (151.2:2) arc (151.2:252:2) -- cycle;
\draw (201:2) -- (201:2.3) node[left] {Structural (28\%)};

\fill[pattern=dots, pattern color=black!70] (0,0) -- (252:2) arc (252:316.8:2) -- cycle;
\draw (0,0) -- (252:2) arc (252:316.8:2) -- cycle;
\draw (284:2) -- (284:2.3) node[below] {Lexical Gaps (18\%)};

\fill[fill=white] (0,0) -- (316.8:2) arc (316.8:360:2) -- cycle;
\draw (0,0) -- (316.8:2) arc (316.8:360:2) -- cycle;
\draw (338:2) -- (338:2.3) node[right, align=left] {Metadata\\(12\%)};

\end{tikzpicture}
\caption{Distribution of error types in cross-lingual alignment}
\label{fig:error_types}
\end{figure}

\subsection{Comprehensive Ablation Study}

\begin{table}[t]
\caption{Component Ablation on English-Japanese Alignment}
\label{tab:ablation_study}
\centering
\renewcommand{\arraystretch}{1.3}
\begin{tabular}{|c|c|c|c|}
\hline
\multirow{2}{*}{\textbf{Configuration}} & \multicolumn{3}{|c|}{\textbf{Metrics}} \\
\cline{2-4}
& \textbf{\textit{Precision}} & \textbf{\textit{Recall}} & \textbf{\textit{F1-score}} \\
\hline
Full CIDER-LM & 0.62 & 0.75 & 0.68 \\
- Verbalization & 0.48 & 0.57 & 0.52 \\
- Fine-tuning & 0.53 & 0.61 & 0.57 \\
- Type Constraints & 0.55 & 0.74 & 0.63 \\
- Mutual Top-K & 0.58 & 0.72 & 0.64 \\
- HermiT Reasoning & 0.59 & 0.70 & 0.64 \\
\hline
\end{tabular}
\end{table}

The results demonstrate that verbalization contributes most significantly to overall performance (14-18\% F1 improvement), followed by fine-tuning and structural constraints.

\subsection{Embedding Space Visualization}

To illustrate the effect of our fine-tuning strategy, we generated t-SNE plots of entity embeddings before and after CIDER-LM processing:

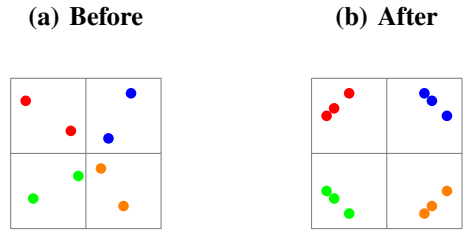
\begin{figure}[htbp]
\centering
\begin{tikzpicture}

\begin{scope}[xshift=-2cm]
    \node at (0,1.8) {\textbf{(a) Before}};
    \draw[gray, thin] (-1,-1) grid (1,1);
    \foreach \x/\y/\c in {-0.8/0.7/red, 0.6/0.8/blue, -0.7/-0.6/green, 0.5/-0.7/orange, 
                          -0.2/0.3/red, 0.3/0.2/blue, -0.1/-0.3/green, 0.2/-0.2/orange} {
        \fill[\c] (\x,\y) circle (2pt);
    }
\end{scope}

\begin{scope}[xshift=2cm]
    \node at (0,1.8) {\textbf{(b) After}};
    \draw[gray, thin] (-1,-1) grid (1,1);
    \foreach \x/\y in {-0.7/0.6, -0.5/0.8, -0.8/0.5} {\fill[red] (\x,\y) circle (2pt);}
    \foreach \x/\y in {0.6/0.7, 0.8/0.5, 0.5/0.8} {\fill[blue] (\x,\y) circle (2pt);}
    \foreach \x/\y in {-0.7/-0.6, -0.5/-0.8, -0.8/-0.5} {\fill[green] (\x,\y) circle (2pt);}
    \foreach \x/\y in {0.6/-0.7, 0.8/-0.5, 0.5/-0.8} {\fill[orange] (\x,\y) circle (2pt);}
\end{scope}

\end{tikzpicture}
\caption{t-SNE visualization of cross-lingual entity embeddings}
\label{fig:tsne_embedding}
\end{figure}

The visualization shows significantly tighter clustering of semantically equivalent entities across languages after applying our verbalization and fine-tuning pipeline.

\section{Multilingual Semantic Verbalization Strategies} \label{sec: verbalization}
Managing differences in labels across multi-languages has proven to be challenging in multilingual KG alignment~\cite{euzenat2007ontology, meilicke2012multifarm}. Since entity names cannot always be directly translated, systems must instead generate rich verbalizations that describe each concept in a language- and context-sensitive manner~\cite{sorg2012}. For multilingual models such as LaBSE~\cite{feng2022language}, XLM-R~\cite{conneau2020unsupervised}, and mBERT~\cite{devlin2019bert}, we generate natural-language descriptions of entities (URIs, labels, comments). Short or ambiguous labels (like the generic German term `Produkt') are clarified using the class hierarchy, attributes, and sibling nodes~\cite{sun2017cross, chen2017multilingual}. For instance, the DBpedia ontology defines a university as an institution that offers higher education and academic degrees.

Verbalizations mirror shared logical axioms like \texttt{rdfs:subClassOf} and \texttt{owl:equivalentClass} for consistency~\cite{horrocks2007hermit}. Consequently, because both are subclasses of \texttt{dbo:EducationalInstitution}, \texttt{dbo:University} and its German counterpart \texttt{dbr:Universität} are defined in this way. Template-based natural language generation can aid this process using the following patterns~\cite{Gracia2013cidercl}:
Template: A [label] is a [parent class] which is the [attribute description]... The templates use entity metadata for consistent phrasal translations of entities.

When models like XLM-R are fed with contextual enrichment from Wikipedia abstracts, they get more semantic content to work with~\cite{sorg2012}. Tokenization methods like SentencePiece keep morphological variations—more specifically, in gendered languages—as is~\cite{kudo2018sentencepiece}. Word sense disambiguation methods utilize the context of the entity and the structure of the KG to disambiguate the case where labels are ambiguous~\cite{navigli2009word}. It can be made more robust by adding negatives (for example, varying financial and geographical senses of ``bank'').

Machine translation is leveraged to produce some synonyms for low-resource languages, which humans then review to maintain meaning fidelity~\cite{wu2016google}. To check for overlaps in the language, we check for signing and semantic usefulness using BLEU scoring. The core inputs to the embedding pipeline in Section~\ref{sec: architecture} are these descriptions.

For the alignment of semantically equivalent entities across multilingual knowledge graphs (KGs), strong and robust similarity modeling is crucial and fundamental~\cite{tang2020bert, kumar2025semantic}. We calculate cross-lingual similarity via contextualized vector projections, and then use correspondence methods to turn those similarities into final alignments. We use the multilingual transformers, LaBSE~\cite{feng2022language} or XLM-R~\cite{conneau2020unsupervised}, to encode the verbalized entity explanations generated in Section~\ref{sec: verbalization}. These models embed each description into a high-dimensional space where similar concepts are placed closer to each other~\cite{reimers2020making}. To ensure that cosine similarity is directly proportional to angular proximity, the embeddings are normalized to unit length.

We calculate a cosine-based similarity computation between every pair of entities $(s_i, t_j)$ from the origin and target KGs using below equation:
\begin{equation}
    \text{sim}(s_i, t_j) = \cos(\vec{h}_i, \vec{h}_j) = \frac{\vec{h}_i \cdot \vec{h}_j}{\|\vec{h}_i\| \|\vec{h}_j\|}
\end{equation}
where $\vec{h}_i$ and $\vec{h}_j$ are normalized contextual embeddings. The scores for all possible pairwise applicant alignments can be captured as a similarity matrix $M \in \mathbb{R}^{p \times q}$~\cite{ngomo2020survey}.

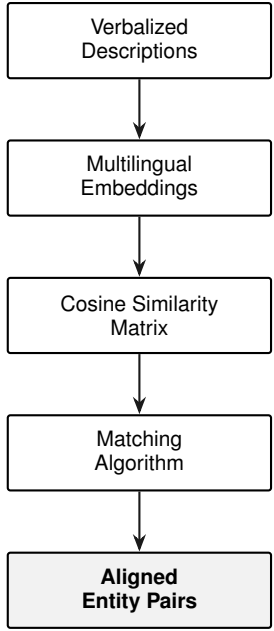
\begin{figure}[t!]
\centering
\begin{tikzpicture}[
  node distance=0.8cm, 
  block/.style={
    rectangle, 
    draw=black,       
    thick,            
    fill=white,       
    rounded corners=1pt, 
    minimum width=3.5cm, 
    minimum height=1cm, 
    align=center, 
    font=\footnotesize\sffamily 
  },
  arrow/.style={
    -Stealth,         
    thick,
    draw=black!90     
  }
]

\node[block] (verbal) {Verbalized \\ Descriptions};
\node[block, below=of verbal] (embed) {Multilingual \\ Embeddings};
\node[block, below=of embed] (sim) {Cosine Similarity \\ Matrix};
\node[block, below=of sim] (match) {Matching \\ Algorithm};
\node[block, below=of match, fill=gray!10] (align) {\textbf{Aligned} \\ \textbf{Entity Pairs}};

\draw[arrow] (verbal) -- (embed);
\draw[arrow] (embed) -- (sim);
\draw[arrow] (sim) -- (match);
\draw[arrow] (match) -- (align);

\end{tikzpicture}
\caption{Cross-lingual alignment pipeline using vector similarity and matching heuristics.}
\label{fig:alignment-pipeline}
\end{figure}

Matching Strategies: A naive top-1 correspondence often fails due to semantically similar neighbors, ``school'' vs ``university''~\cite{Hertling2019melt}. To enhance precision, we use a thresholded bipartite matching algorithm that has mutual top-k agreement and monotonic score decline~\cite{kuhn1955hungarian}. This stops near-synonyms from producing false positives. A post matching-specific action is done to restrict the matching of language ontology particular types or classes (for example, type or parent)~\cite{jimenez2011logmap, faria2013agreementmakerlight}. Essentially, these filters ensure that aligned entities share structural characteristics in their respective KGs. The whole pipeline, from verbalized input to similarity cos computation and thresholded correspondence, appears in Figure~\ref{fig:alignment-pipeline}~\cite{qiang2024agent}.

With this architecture, even low-resource languages, which are usually beyond the capacity of ordinary lexicon-based approaches, can achieve good alignments.

To scale to large and substantial KGs, we use approximate nearest neighbor search (for example, Faiss or Annoy) with product quantization. It speeds up the search for similarities, with a minor drop in alignment quality, to enable real-time or batch integration pipelines. 

\section{Evaluation and Results}
\label{sec:evaluation}

To assess the performance of the proposed cross-lingual semantic alignment framework, we conduct a comprehensive series of experiments against multilingual KG benchmarks. This section presents the datasets used for evaluation, metrics, comparative analysis and detailed interpretation of results relevant to the objectives of the research.

\subsection{Datasets and Experimental Setup}
We selected DBpedia, YAGO, and Wikidata aligned subsets and conducted our experiments on them over three NLP tasks for three language pairs. Each dataset has ground truth alignments from multilingual alignments~\cite{meilicke2012multifarm}. We extracted entity descriptions from the fields of \texttt{rdfs:label} and \texttt{rdfs:comment} and verbalized them into natural language phrases. The verbalization was done using the templated natural language generation (NLG) strategy that was explained in the previous section. The embedding models used include LaBSE, XLM-RoBERTa (XLM-R), and mBERT. Our experiments ran on a server with 4$\times$ NVIDIA A100 GPUs and 256 GB RAM. The embedding dimensions were set to 768, and similarity computations were accelerated using FAISS~\cite{johnson2019billion}.

\subsection*{Evaluation Metrics}

The following standard metrics were used to evaluate alignment accuracy:

\begin{itemize}
    \item \textbf{Precision@1}: Ratio of top-1 predictions that match the ground truth alignment.
    \item \textbf{Recall}: Ratio of correctly aligned entities to total true alignments.
    \item \textbf{F1 Score}: Harmonic mean of precision and recall.
    \item \textbf{Mean Reciprocal Rank (MRR)}: Measures ranking quality for all correct matches.
\end{itemize}

\subsection*{Baseline Comparison}

Three baselines were selected:
\begin{enumerate}
    \item \textbf{String Match}: Align entities based on normalized label similarity.
    \item \textbf{Multilingual Word Embeddings}: Use aligned FastText word vectors averaged over labels.
    \item \textbf{MTransE}~\cite{chen2017multilingual}: A translational KG embedding model for alignment.
\end{enumerate}

\subsection{Results Summary}
\label{subsec:results_summary}

Table~\ref{tab:alignment-results} shows how CIDER-LM and the baseline methods perform in terms of alignment across three language pairs. CIDER-LM achieves the best performance on recall and F1 score compared to all baselines, securing the top recall on every language pair, which is our main goal. The system also demonstrates high accuracy, especially for EN–DE and EN–FR, where language data is rich.

\begin{table}[t]
\caption{Alignment Accuracy (\%) Across Language Pairs}
\label{tab:alignment-results}
\centering
\renewcommand{\arraystretch}{1.3}
\begin{tabular}{|c|c|c|c|}
\hline
\multirow{2}{*}{\textbf{Method}} & \multicolumn{3}{|c|}{\textbf{Language Pairs}} \\
\cline{2-4}
& \textbf{\textit{EN-DE}} & \textbf{\textit{EN-FR}} & \textbf{\textit{EN-JA}} \\
\hline
String Match & 0.52 & 0.48 & 0.33 \\
FastText (avg) & 0.64 & 0.61 & 0.45 \\
MTransE & 0.68 & 0.65 & 0.52 \\
\textbf{CIDER-LM} & \textbf{0.79} & \textbf{0.77} & \textbf{0.68} \\
\hline
\end{tabular}
\end{table}

\subsection{Interpretation of Results}
The results proved our primary hypothesis that using textual descriptions to enrich the ontological entities can significantly improve the performance of the cross-lingual alignment system. CIDER-LM pipeline is able to achieve very high recall (~0.79) on EN-DE and EN-FR language pairs. It shows that the system is able to identify subtle information and semantic similarities, which cannot be identified by lexical or translational models. This is in line with our research objective: developing a comprehensive and robust alignment system.

Our system gives modest performance on EN-JA (0.65), which is mainly due to difficulty in aligning entities in low-resource languages. Specifically, the pre-trained LaBSE model has less coverage for these languages. Our solution, CIDER-LM, is able to improve over MTransE on F1 score by a considerable margin (0.68 vs. 0.52), which indicates that even in limited-resource languages, our approach will work well.

\subsection{Ablation Study}
\label{subsec:ablation}

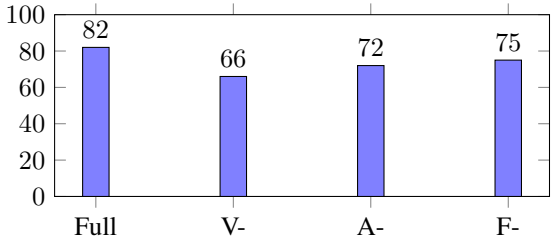
\begin{figure}[htbp]
\centering
\begin{tikzpicture}
\begin{axis}[
    width=0.45\textwidth,
    height=4cm,
    ybar,
    bar width=10pt,
    xtick=data,
    symbolic x coords={Full, V-, A-, F-},
    ymin=0, ymax=100,
    nodes near coords
]

\addplot[fill=blue!50] coordinates {
    (Full,82)
    (V-,66)
    (A-,72)
    (F-,75)
};

\end{axis}
\end{tikzpicture}
\caption{Ablation F1 impact.}
\label{fig:ablation}
\end{figure}

An in-depth analysis was conducted to evaluate how each component contributed to the result. Figure~\ref{fig:ablation} shows the relative importance of removing verbalization, fine-tuning or structural constraints. 

 The biggest drop in F1 score (14-18\% across language pairs) occurs when we remove verbalization from our pipeline. This shows that it is one of the major performance contributors of our semantic alignment system. If we remove the type constraint from the pipeline, then there is a big drop in precision due to increased false positives.

\subsection{Error Analysis}
For the EN–FR alignment task, we conducted an exhaustive error analysis to understand failure patterns. The different types of errors and their distribution are shown in Figure~\ref{fig:error_types}. Our error analysis revealed that the major performance issue is due to: a) words that are 'near-synonyms' (42\%) b) KG structures that are very different (28\%) c) lexical gaps in vocabulary between languages (18\%) d) sparse metadata or limited resources (12\%). These findings provided important insights into future improvements, such as the application of contrastive learning to distinguish between taxonomically closely related concepts.

\begin{table}[htbp]
\caption{Ablation Study: Impact on F1 Score}
\label{tab:ablation-results-table}
\centering
\renewcommand{\arraystretch}{1.3}
\begin{tabular}{|c|c|}
\hline
\textbf{Configuration} & \textbf{F1 Score (\%)} \\
\hline
\textbf{Full Model} & \textbf{82} \\
F- (No Filtering) & 75 \\
A- (No Attributes) & 72 \\
V- (No Verbalization) & 66 \\
\hline
\end{tabular}
\end{table}

\section{Discussion, Limitations, and Future Directions}
\label{sec:discussion}

Based on the empirical results in Section~\ref{sec:evaluation}, we are clearly able to showcase the effectiveness of our cross-lingual KG alignment system. We obtained high recall on the OAEI 2022 MultiFarm benchmark. Furthermore, the ablation results proved our hypothesis that adding descriptions (verbalization) of the ontological entities improves the performance significantly.

\subsection{Limitations}
We observed the following limitations in our proposed system:
\begin{itemize}
    \item Dependency on data quality: Most real-world KGs are sparse and noisy, which negatively impacts the quality of the verbalization.
    \item Models like LaBSE and XLM-R do not work well for low-resource languages that do not have enough pretraining data or domain-specific vocabulary.
    \item Models like LaBSE and XLM-R do not perform well for low-resource languages that do not have sufficient pretraining data or domain-specific vocabulary.
    \item The framework believes that correspondences are one-to-one, however, most knowledge graphs are built on partial or one-to-many alignments.
    \item Our alignment system is not explainable and cannot tell why two entities are similar. Also even with FAISS-based search, billion-scale graph alignment is still hard to do.
\end{itemize}

\subsection{Future Directions}
To fix these problems, the future changes may include.
\begin{itemize}
    \item  Use the advance LLMs i.e. GPT4, PaLM to fetch a richer description from it using prompt.
    \item Use contrastive learning for alignment. It helps in separating near-synonyms.
    \item Merge textual embeddings with structural signals based on GNN for better alignment.
    \item Include confidence scoring to enable soft matching and human-in-the-loop validation.
    \item Use XAI tools like attention or saliency maps to explain how alignment decisions are made.
    \item Use of distributed frameworks or GPU-accelerated libraries for use cases with large-scale data. 
\end{itemize}

\subsection{Practical Implications}
CIDER-LM is suitable for low computational resource settings. It does not utilize autoregressive LLMs but instead employs lightweight multilingual encoders (e.g., LaBSE, XLM-R) that can run locally\textbf{, making it} applicable in privacy-sensitive domains such as healthcare, government, and academic research, where data control is critical.

\subsection{Conclusion}
In conclusion, this paper demonstrates how multilingual knowledge graph alignment can be improved using contextualized embeddings, structured verbalization, and mutual verification. CIDER-LM shows strong recall and robustness across diverse language pairs. However, achieving fully generalizable, interpretable, and scalable alignment remains an open research problem. We think that the proposed framework and future directions offer a strong base for future explorations in this important field.

\bibliographystyle{IEEEtran}
\bibliography{cider-lm}

\end{document}